\documentclass[runningheads]{llncs}
\usepackage{graphicx}
\usepackage{xcolor}
\usepackage{booktabs}
\usepackage{tabularx}
\usepackage[hyperindex,breaklinks]{hyperref}
\usepackage{algorithm} 
\usepackage{algpseudocode}
\usepackage{refcount}
\usepackage{soul}
\usepackage[misc]{ifsym}

\begin{document}
\title{\emph{StatMix}: Data augmentation method that relies on image statistics in federated learning}
\titlerunning{\emph{StatMix}: Data augmentation method for federated learning}

\author{Dominik Lewy\inst{1}\Letter\orcidID{0000-0003-2107-4909} \and
Jacek Ma{\'n}dziuk\inst{1}\orcidID{0000-0003-0947-028X} \and
Maria Ganzha\inst{1}\orcidID{0000-0001-7714-4844} \and
Marcin Paprzycki\inst{2,3}\orcidID{0000-0002-8069-2152}}
\authorrunning{D. Lewy, et al.}

\institute{Faculty of Mathematics and Information Science, Warsaw University of Technology, Koszykowa 75, 00-662 Warszawa, Poland\\
\email{d.lewy@mini.pw.edu.pl, \{jacek.mandziuk,maria.ganzha\}@pw.edu.pl}
\and
Systems Research Institute Polish Academy of Sciences\\
\and 
Warsaw Management University, Warsaw, Poland\\
\email{marcin.paprzycki@ibspan.waw.pl}
}

\maketitle 

\begin{abstract}
Availability of large amount of annotated data is one of the pillars of deep learning success. Although numerous big datasets have been made available for research, this is often not the case in real life applications (e.g. companies are not able to share data due to GDPR or concerns related to intellectual property rights protection). Federated learning (FL) is a potential solution to this problem, as it enables training a global model on data scattered across multiple nodes, without sharing local data itself. However, even FL methods pose a threat to data privacy, if not handled properly. Therefore, we propose \emph{StatMix}, an augmentation approach that uses image statistics, to improve results of FL scenario(s). \emph{StatMix} is empirically tested on CIFAR-10 and CIFAR-100, using two neural network architectures. In all FL experiments, application of \emph{StatMix} improves the average accuracy, compared to the baseline training (with no use of \emph{StatMix}). Some improvement can also be observed in non-FL setups.

\end{abstract}

\section{Introductions}
One of key factors, behind the success of deep learning in Computer Vision, is the availability of large annotated datasets like ImageNet~\cite{IMAGENET} or COCO~\cite{COCO}. However, even if large datasets theoretically exist, there can be restrictions related to bringing them to one place, to enable model training. Federated learning (FL) addresses this challenge by enabling data to be kept where it is, and share only limited information, based on which the original content cannot be recreated. At the same time FL allows training a model that achieves better results than ones trained in isolation on separated nodes. This, for instance, is a typical scenario for hospitals that gather (possibly annotated) medical images. However, they cannot share it with other hospitals, due to various reasons (e.g. GDPR regulations or intellectual property rights protection). 

According to the FL classification, proposed in~\cite{Survey_2019}, the method presented in this paper addresses a horizontal data partitioning scenario (each of individual nodes collects similar data). The specific focus is on Convolutional Neural Network (CNN) architectures, since the problem considered is an image classification. However, the approach is in no way limited to the CNN-oriented use case. The proposed method is based on sharing limited amount of data between nodes, thus avoiding violation of privacy. In the paper we consider a centralized FL setup. Nevertheless, the proposed algorithm (\emph{StatMix}) is communication architecture agnostic and can be easily applied in decentralized settings with each node sharing information with all the other (or selected group of) nodes, instead of a server. Again, with assumed minimization of amount of shared information, the efficiency of communication is not the focus of this study.

\subsection{Motivation}
Historically, in majority of FL research, during model training, either gradients of the training process (e.g. \emph{FedSGD}~\cite{FedAvg}) or weights of the model (e.g. \emph{FedAvg}~\cite{FedAvg}), have been shared. Only recently a paper on sharing averaged images (\emph{FedMix}~\cite{FedMix}) was published. However, all these approaches pose a potential threat to data privacy if data sharing is not properly managed (e.g. by using differential privacy, or by ensuring the number of images in the averaged images is large enough). The method, proposed in what follows, limits the information shared to bare minimum (just 6 values, 2 per each color channel), and is still able to provide boost in accuracy. 

\subsection{Contribution}
The main contribution of this work is threefold:
\begin{itemize}
    \item A simplistic data augmentation (DA) mechanism (\emph{StatMix}), dedicated to FL learning setup that limits the amount of communication between participating nodes, is proposed.
    \item \emph{StatMix} is evaluated on two different CNNs, with numbers of FL participants ranging from 5 to 50.
    \item It is shown that the standard set of simple DAs, typically used for CIFAR datasets, is not well suited for FL scenario, as it deteriorates the performance along with a decrease of the number of samples per each FL node.
\end{itemize}

\section{Related work}
\subsubsection{Federated learning} 

Since FL system is, usually, a combination of algorithms each research contribution can be regarded and analysed from different angles. Typical FL aspects include: (1) if the data is partitioned horizontally or vertically~\cite{FedLearnVertical}, (2) which models are used (some require dedicated algorithms, e.g. trees~\cite{FedLearnTrees}, other can be addressed with more general methods, like SGD~\cite{SGD}), 
(3) whether the global model is updated during the training process~\cite{FedAvg}, or only once when all local models have been already trained~\cite{FedLearnNNBlending}, (4) what (if any) is the mechanism that guarantees privacy of the data~\cite{FedLearnDifferentialPrivacy}, and/or (5) how effective is the process of sharing information between parties of the system~\cite{FedLearnEfficiency}.

The idea of FL was introduced in~\cite{FedLearnCanonical}, where the usage of asynchronous SGD to update a global model in a distributed fashion was proposed. Currently, the most common approach is~\emph{FedAvg}~\cite{FedAvg}, which at each communication round, performs training on a fraction of nodes, using the local data and, at the end of each round, averages the model weights on the server.
Subsequent works, in this area, focused on either making the process more effective~\cite{FedLearnEfficiency,FedLearnEfficiency2}, or being able to address particular data-related scenarios (e.g. non-IID setup~\cite{FedProx,FedLearnNonIID,non-IID}). Since the \emph{StatMix} method shares only highly limited information between nodes, due to space limits, privacy guarantees and communication efficiency will not be discussed in the literature review.

\subsubsection{Data Augmentation} 
Another research area relevant to the scope of this paper is DA~\cite{naszeSurvey}, especially the methods dedicated to the FL setup. An interesting research approach is adjustment of \emph{Mixup}~\cite{mixup} to the FL regime (\cite{FedMix,FedLearnMixupUtilization,FedLearnMixupUtilization2}). However, it requires sending mixed data to the server rendering these methods expensive in terms of communication. Moreover, in some cases, this could lead to privacy violation, if small number of samples is selected for mixing.

An alternative approach to DA, is the use of GANs for local node DA~\cite{GANFederatedAugmentation,GANFederatedAugmentationv2}. These approaches require samples from private node data, to be shared with the server for the purpose of GAN training that will be subsequently downloaded to each node to generate additional synthetic samples.

Another approach to synthetic data generation is the usage of models trained using, for instance, \emph{FedAvg}, to generate samples based on the statistics from the batch normalization layer, using a Zero Shot Learning~\cite{ZSLFederatedAugmentation}.

Yet another stream of research, worth mentioning that according to our best knowledge was not yet applied to FL problems, and is an inspiration for this work, is 
\emph{MixStyle}~\cite{mix_style}, which is dedicated to the problem of Domain Generalization (DG), i.e. construction of classifiers robust to domain shift, able to generalize to unseen domains. To this end \emph{MixStyle}, similarly to \emph{Mixup} based methods, performs sample mixing, However, it does not mix pixels but instance-level feature statistics of the two images generated from the neural network.

\section{Proposed approach }
\label{sec:approach}

\begin{algorithm}[ht!]
    \textbf{\textit{Local part 1:}}
    \begin{algorithmic}[1]
        \State $K \leftarrow$ number of images in the node; $N \leftarrow$ number of nodes
		\For {$i=1,2,\ldots, N$}
    		\For {$k=1,2,\ldots, K$}
    				\State Calculate all the image statistics according to equations (\ref{eq:mean})-(\ref{eq:stdev})
    				\State $S_{ik} = \{\mu(x_{ik})_{1}, \mu(x_{ik})_{2}, \mu(x_{ik})_{3}, \sigma(x_{ik})_{1}, \sigma(x_{ik})_{2}, \sigma(x_{ik})_{3}\}$
    		\EndFor
    	\EndFor
		\State Share statistics with the sever
    \end{algorithmic}
    \vspace{8pt}
	\textbf{\textit{Sever part:}}
	\begin{algorithmic}[1]
	\setcounter{ALG@line}{8}
	\State Distribute statistics to all nodes
	\end{algorithmic}
	\vspace{8pt}
	\textbf{\textit{Local part 2:}}
	\begin{algorithmic}[1]
	\setcounter{ALG@line}{9}
\For {$i=1,2,\ldots, N$}
	\For {$epoch=1,2,\ldots, max\_epoch$}
    	\For {$batch=1,2,\ldots, max\_batch$}
    	    \If {$random(0,1) < P_{StatMix}$}
    		\State Randomly select set of statistics $S_{jm}, j\in\{1,\ldots,N\}, m\in\{1,\ldots,K\}$ 
    		\State Normalize images from a batch using equation (\ref{eq:norm}) 
    		\State Apply augmentation using equation (\ref{eq:augment}) 
    		\EndIf
    	\EndFor
	\EndFor
\EndFor
	\end{algorithmic}
\caption{\emph{StatMix}} 	
\label{alg:1}
\end{algorithm}

In a typical FL scenario, there are two main components: nodes which contain local data that cannot be shared (e.g., due to privacy reasons), and a server that coordinates the process of information exchange. In certain FL implementations the central server is not used, and participating nodes communicate directly.

\begin{figure}[ht!]
  \includegraphics[width=\linewidth]{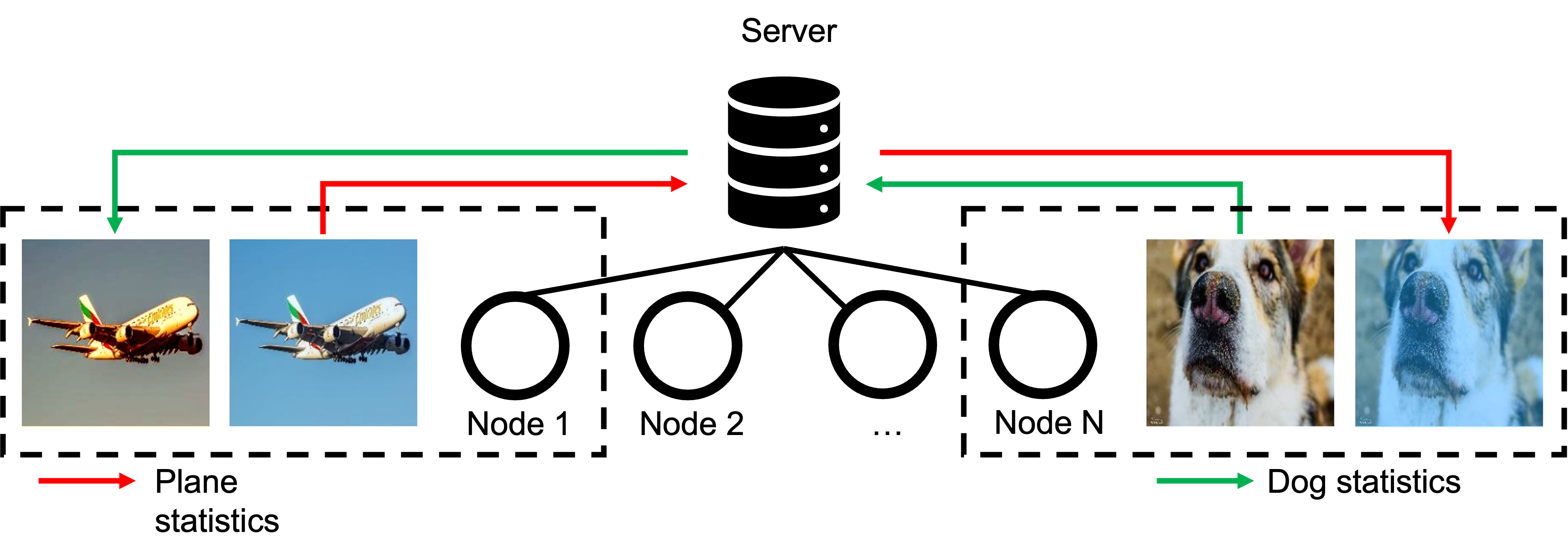}
  \caption{The figure is composed of two components: a central server (storing only image statistics) and nodes (storing subsets of images and image statistics obtained from the server). The flow shows application of statistics calculated in one node to augment images in another node. For instance, node 1 shares statistics of a plane image with node N, based on which an augmented image of a dog is created.}
  \label{fig:concept}
\end{figure}

The goal of this work is to increase the accuracy of classifiers, trained in individual nodes, 
by using limited statistical information (delivered by all nodes and aggregated on the server). This is to be achieved without sending/storing any actual data. Overall, the proposed approach can be characterized as follows (see Figure~\ref{fig:concept} and Algorithm~\ref{alg:1}):
\begin{itemize}
    \item[(a)] Calculation of image statistics (mean and standard deviation per color channel) in individual nodes -- \textit{Local part 1} in Algorithm~\ref{alg:1} 
    \item [(b)] Distribution of the calculated statistics to all nodes via central server -- \textit{Server part} in Algorithm~\ref{alg:1}
    \item [(c)] Using these statistics in individual nodes to perform style transfer like augmentation of images in this node -- \textit{Local part 2} in Algorithm~\ref{alg:1}.
\end{itemize}

\begin{figure}[ht!] 
  \includegraphics[width=\linewidth]{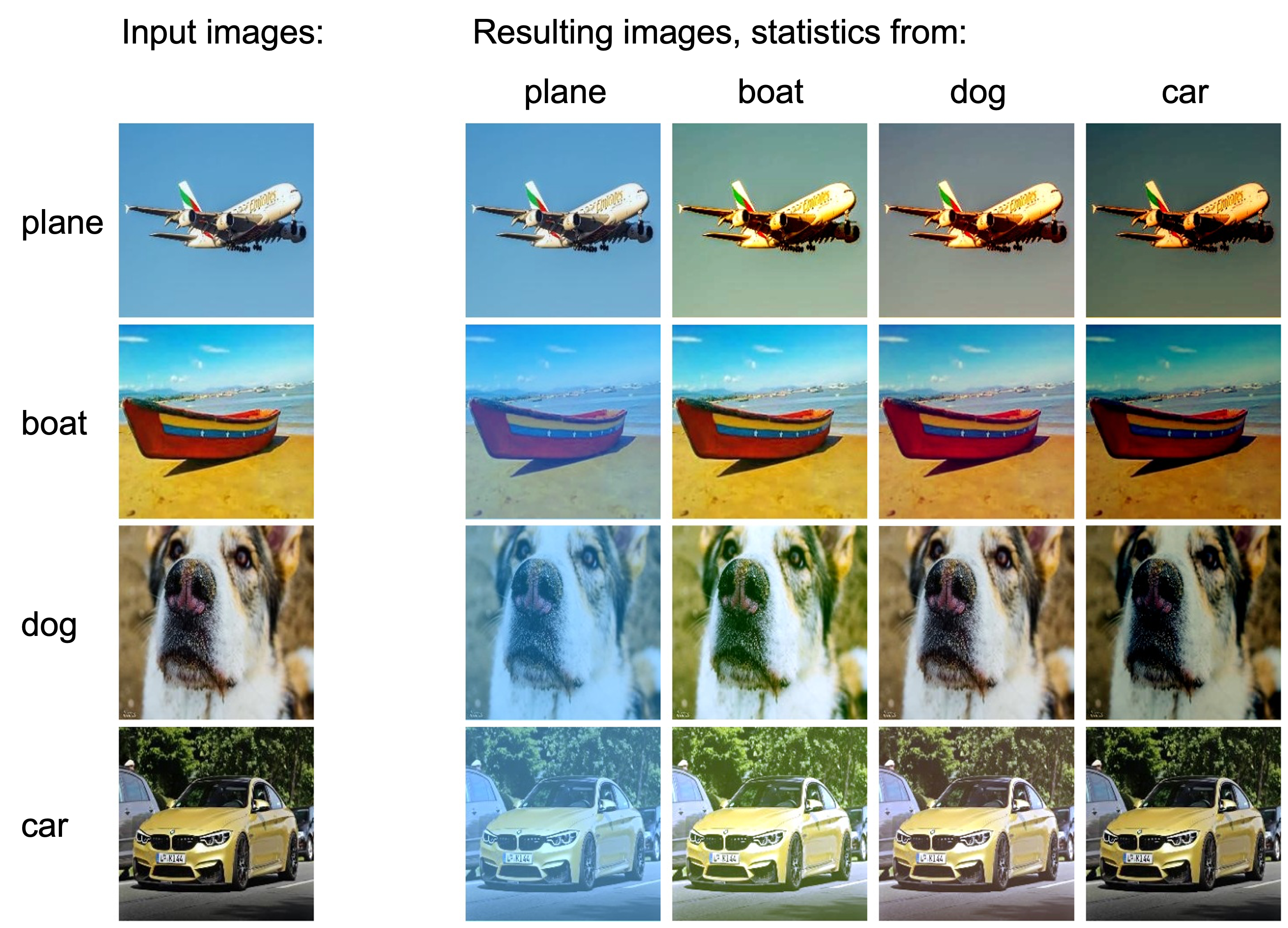}
  \caption{The first column shows original images and the remaining part of the figure depicts these images augmented with statistics of various images (each column utilizes a different set of image statistics).}
  \label{fig:example}
\end{figure}

\subsubsection
{\textit{Local part 1.}} This is the first step of the algorithm. In each node $i=1,\ldots,N$, for each locally stored image $x_{ik}$, $k=1,\ldots,K$, where $x_{ik} \in R^{W \times H \times C}$ ($W, H$ and $C$ denote width, height and color channel, resp.), the mean and the standard deviation of image pixels are calculated separately for each colour channel $C=1,2,3$, using the following equations:
\begin{equation}
    \mu(x_{ik})_{c} = \frac{1}{HW} \sum_{h=1}^{H} \sum_{w=1}^{W} x_{ik}[w,h,c]
\label{eq:mean}
\end{equation}
\begin{equation}  
\sigma(x_{ik})_{c} = \sqrt{\frac{1}{HW} \sum_{h=1}^{H} \sum_{w=1}^{W} \Bigl(x_{ik}[w,h,c] - \mu(x_{ik})_{c}\Bigr)^{2} }
\label{eq:stdev}
\end{equation}
where $x_{ik}[w,h,c]$ is a value of $[w,h]$ pixel of image $x_{ik}$, in color channel $c$. These $6$ statistics form the set $S_{ik}$, that is used in \emph{Local part 2} for image augmentation.

\subsubsection
{\textit{Server part.}} In the second step, all sets $S_{ik}$, $i=1,\ldots,N$, $k=1,\ldots,K$, are distributed to all $N$ nodes, i.e. in each node, in addition to $K$ local (private) images, $N\cdot K$ statistics are now stored. 
\subsubsection{\textit{Local part 2.}} 
Next the augmentation part takes place. In each node $i=1,\ldots,N$, all images $x_{ik}$ located in that node ($k=1,\ldots,K$) are randomly divided into $max\_batch$ batches. Then, for each batch, an image $x_{jm}$ is uniformly selected (from all $N\cdot K$ images, i.e. including those located in a given node
) and the corresponding set of statistics $S_{jm}$ is applied to augment all images from the batch 
using equations (\ref{eq:norm})-(\ref{eq:augment}). This augmentation procedure is applied independently to each batch with probability $P_{StatMix}$.
\begin{equation}
    x^{norm}_{ik,c} = \frac{x_{ik,c} - \mu(x_{ik})_{c}}{\sigma(x_{ik})_{c}}  
\label{eq:norm}
\end{equation}
\begin{equation}
x^{augment}_{ik,c} = x^{norm}_{ik,c} \cdot \sigma(x_{jm})_{c} + \mu(x_{jm})_{c}
\label{eq:augment}
\end{equation}

Note that augmentation procedure (\ref{eq:norm})-(\ref{eq:augment}) is applied independently to all $3$ color channels. 
Example results of \emph{StatMix} augmentation are depicted in Figure~\ref{fig:example}.

\section{Experimental setup}
The experiments were conducted with two popular datasets.
\textbf{CIFAR-10}~\cite{cifar_10} consists of $50\,000$ training and $10\,000$ test color images, of size $32 \times 32$, grouped into $10$ classes (airplane, automobile, bird, cat, deer, dog, frog, horse, ship and truck). There are $5\,000$ and $1\,000$ samples of each class in the training and test datasets, respectively.  

\textbf{CIFAR-100}~\cite{cifar_100} is a more granular version of CIFAR-10, with 100 classes. Each class has $500$ representatives in the training and $100$ in the test datasets, respectively. 

In order to simulate the FL scenario, let us denote by $P$ the set of all training images in a given dataset (CIFAR-10, or CIFAR-100, respectively). $P$ was randomly divided, in a stratified manner, into $N$ disjoint subsets ($P_1,\ldots,P_N$) of equal size, using labels to reflect the same distribution in each $P_i$, as in the whole set $P$. Subsequently, each part was transferred (assigned) to a separate FL node that was connected only to the server (i.e. there were no connections between FL nodes). At this point each of $N$ nodes calculated statistics of images located in this node and transferred them on the server. 
Next, for each node $i=1,\ldots,N$, the server shared 
individual statistics of all images not located in node $i$, i.e. all images from $P \setminus P_i$. 
Based on this, the images located in node $i$, i.e. those belonging to $P_i$, could 
be augmented (with certain probability) using image statistics from the entire data set $P$, according to the approach described in section~\ref{sec:approach}. The augmented sets $PA_i, i=1,\ldots,N$ were used to train the model (one of the $2$ deep architectures described in the following paragraph). Afterwards, the trained model was tested on the entire test part of the respective CIFAR dataset.

Two popular \textbf{architectures} were tested during experiments: PreActResNet18~\cite{PreActResNet18} and DLA~\cite{DLA}. The models belong to different families and offer decent accuracy in non-FL scenarios.

SGD optimizer with initial learning rate equal to $0.01$ and momentum equal to $0.9$ was used. The learning rate was adapted, using cosine annealing~\cite{cosineannealing}, from the initial learning rate to $0$, over the course of the training process. In all experiments that mention standard DA, random image crop and random horizontal flip were applied~\cite{AlexNet}. 
For consistency, all models were trained for 200 epochs, on a batch of 128 images at a time.

The experiments were ran 3 times for each $N = 1, 5, 10, 50$ with the probability of applying statistics-based augmentation set to $0.5$. 

\section{Experimental results and analysis}

First, \textbf{CIFAR-10} results are presented in Table~\ref{tab:cifar10}.
\begin{table}[h]
\caption{
Mean and standard deviation results for CIFAR-10 dataset averaged over last $10$ epochs and 3 experiment repetitions. Columns denote: number of nodes ($N$), model architecture, whether or not standard DA was applied, whether \emph{StatMix} augmentation was used ($0.0$ -- not used, $0.5$ -- used with probability $0.5$), the relative improvement of applying \emph{StatMix} compared to not applying it, i.e. [mean(${0.5}$) / mean(${0.0}$) -- 1].
}\label{tab:cifar10}
\begin{tabularx}{\textwidth}{c|l|l|X|X|X|X|X}
\toprule
   &                &      & \multicolumn{4}{c}{\emph{StatMix}} & \\
   &                &  & \multicolumn{2}{c}{0.0} & \multicolumn{2}{c}{0.5} &  \\
Nodes ($N$) & Architecture & Standard &     mean &   std &   mean & std & diff [\%]\\
\midrule
1  & DLA & False &   86.02 &  0.80 &  86.58 &  0.47 &     0.65 \\
   &                & True &   93.26 &  0.28 &  93.83 &  0.19 &     0.61 \\
   & PreActResNet18 & False &   86.15 &  0.79 &  86.60 &  0.14 &     0.52 \\
   &                & True &   93.54 &  0.05 &  93.79 &  0.13 &     0.27 \\
   \midrule
5  & DLA & False &   67.32 &  1.15 &  69.47 &  0.70 &     3.19 \\
   &                & True &   63.39 &  1.03 &  66.24 &  0.89 &     4.50 \\
   & PreActResNet18 & False &   70.83 &  0.44 &  72.01 &  0.55 &     1.67 \\
   &                & True &   68.22 &  0.64 &  69.12 &  0.33 &     1.32 \\
   \midrule
10 & DLA & False &   56.06 &  1.27 &  58.97 &  1.09 &     5.19 \\
   &                & True &   50.72 &  1.45 &  54.54 &  1.59 &     7.53 \\
   & PreActResNet18 & False &   60.72 &  0.64 &  62.03 &  0.76 &     2.16 \\
   &                & True &   56.63 &  0.77 &  58.69 &  0.74 &     3.64 \\
   \midrule
50 & DLA & False &   37.47 &  1.20 &  38.06 &  1.42 &     1.57 \\
   &                & True &   34.06 &  1.11 &  34.65 &  1.39 &     1.73 \\
   & PreActResNet18 & False &   38.62 &  0.96 &  40.28 &  1.08 &     4.30 \\
   &                & True &   35.01 &  1.07 &  36.93 &  1.21 &     5.48 \\
\bottomrule
\end{tabularx}
\end{table}
In all experiments, in the FL setup ($N > 1$) the application of \emph{StatMix} boosts the final accuracy, compared to the baseline case, with no use of \emph{StatMix}. The impact of the method grows with the number of nodes in the system (at least, to 50 nodes, as tested here). 

It is worth noting that the augmentation method, proposed for the FL setup, works also in a non-FL scenario ($N=1$). The improvement can be observed in all 4 cases (cf. column \emph{diff [\%]} in Table~\ref{tab:cifar10}).
Lastly, it can be observed that standard DAs (random crop and horizontal flip), often utilized with CIFAR data in non-FL scenarios, deteriorate the accuracy of training in the FL scenario (cf. row \emph{True} vs. row \emph{False} for a given architecture and given $N>1$). 

For the \textbf{CIFAR-100}, results are summarized in Table~\ref{tab:cifar100}.
\begin{table}[h]
\caption{
Mean and standard deviation results for CIFAR-100 dataset, average from $10$ epochs and 3 experiment repetitions. Columns, denote: number of nodes ($N$), model architecture, whether or not standard DA was applied, whether \emph{StatMix} augmentation was used ($0.0$ -- not used, $0.5$ -- used with probability $0.5$), relative improvement of applying \emph{StatMix} compared to not applying it, i.e. [mean(${0.5}$) / mean(${0.0}$) -- 1].
}\label{tab:cifar100}
\begin{tabularx}{\textwidth}{c|l|l|X|X|X|X|X}
\toprule
   &                &      & \multicolumn{4}{c}{\emph{StatMix}} & \\
   &                &  & \multicolumn{2}{c}{0.0} & \multicolumn{2}{c}{0.5} &  \\
Nodes ($N$) & Architecture & Standard &     mean &   std &   mean & std & diff [\%]\\
\midrule
1  & DLA & False &   59.29 &  2.08 &  58.11 &  0.87 &    -1.99 \\
   &                & True &   73.40 &  0.26 &  75.25 &  0.46 &     2.52 \\
   & PreActResNet18 & False &   54.99 &  2.73 &  55.84 &  2.21 &     1.55 \\
   &                & True &   71.83 &  0.49 &  73.63 &  0.22 &     2.51 \\
   \midrule
5  & DLA & False &   26.46 &  0.49 &  28.04 &  0.53 &     5.97 \\
   &                & True &   22.84 &  0.71 &  24.84 &  0.60 &     8.76 \\
   & PreActResNet18 & False &   31.02 &  0.58 &  31.39 &  0.58 &     1.19 \\
   &                & True &   27.70 &  0.60 &  28.63 &  0.59 &     3.36 \\
   \midrule
10 & DLA & False &   19.86 &  0.59 &  20.49 &  0.66 &     3.17 \\
   &                & True &   16.48 &  0.57 &  17.80 &  0.92 &     8.01 \\
   & PreActResNet18 & False &   22.32 &  0.41 &  22.86 &  0.50 &     2.42 \\
   &                & True &   19.37 &  0.50 &  20.33 &  0.57 &     4.96 \\
   \midrule
50 & DLA & False &    9.65 &  0.64 &   9.56 &  0.72 &    -0.93 \\
   &                & True &    7.83 &  0.69 &   7.77 &  0.74 &    -0.77 \\
   & PreActResNet18 & False &   10.74 &  0.46 &  10.48 &  0.56 &    -2.42 \\
   &                & True &    9.15 &  0.45 &   9.20 &  0.48 &     0.55 \\
\bottomrule
\end{tabularx}
\end{table}
On this, more granular, dataset similar observations are also valid. In the majority of cases, application of \emph{StatMix} improves the results, compared to the baseline (i.e. the case with no \emph{StatMix} utilization). However, for this more fine-grained dataset this conclusion does not reach 50 nodes, as for this setup, adding \emph{StatMix} deteriorates the performance. This is, most probably, caused by too high noise-to-image ratio after augmentation, due to 10 times smaller number of representatives in individual classes, as compared to CIFAR-10.

The conclusion that \emph{StatMix} is generally beneficial in non-FL scenarios ($N=1$) is also valid for CIFAR-100 (cf. the rightmost column in the table). In 3 out of 4 cases (including both with standard DA application), adding \emph{StatMix} augmentation improves obtained results.

Observation, that standard augmentation methods, used commonly in the literature, do not help in the FL scenario(s), holds also for CIFAR-100. The reason behind that might be that augmentation introduces some noise and the network cannot distill true patterns based on limited amount of clean data.

\subsection{Ablation study}

In order to check how the probability of applying \emph{StatMix} impacts final classification accuracy, additional experiments were performed using PreActResNet18 architecture (which is less computationally intensive than DLA) and a setup with 5 nodes ($N=5$). 
All remaining training parameters were adopted from the base experiments. Probabilities ranging from $0$ to $1$, with a step of $0.1$, were tested. 

\begin{figure}[h]
\center
    \includegraphics[width=0.49\textwidth]{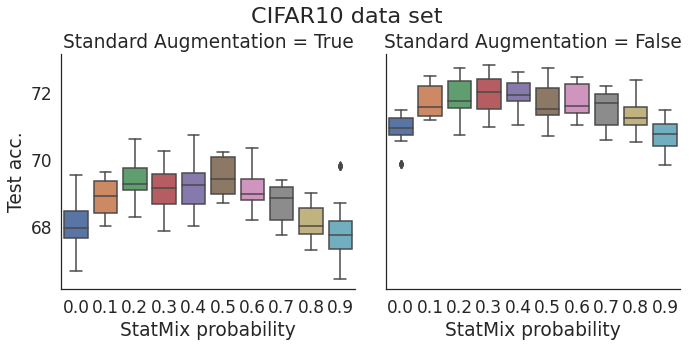}
    \includegraphics[width=0.49\textwidth]{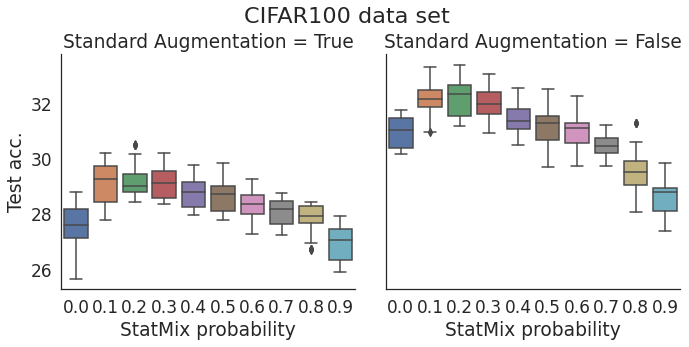}
    \caption{
        CIFAR10 and CIFAR100 test accuracy as a function of probability of applying \emph{StatMix} in FL setup with 5 nodes ($N=5$) on PreActResNet18 architecture. 
    The values are averaged over last $10$ epochs and 3 independent experiment repetitions.
    For each dataset the left figure refers to experiments that utilize standard input DA, the right one presents results without its application.
    }
    \label{cifar10_100:ablation}
\end{figure}

The results for CIFAR10 and CIFAR100 are presented in Figure~\ref{cifar10_100:ablation}. It can be concluded from the chart that both not applying \emph{StatMix} at all, as well as applying it to the majority of the batches (more than $80\%$ for CIFAR10 and more than $60\%$ for CIFAR100) renders the worst results. 

Quite interestingly, applying \emph{StatMix} to all batches ($P_{StatMix}=1$) results in a huge accuracy deterioration (for CIFAR10 the accuracy dropped to $63\%$, while for CIFAR100 to $19\%$ in comparison to no augmentation). These results have been excluded from the charts, to avoid obfuscating other findings. 

For CIFAR10, all probabilities between $0.1$ and $0.8$ bring positive impact, however with no clearly best values. Hence, as long as \emph{StatMix} is applied to a certain fraction of the batches, it leads to accuracy boost. 
For CIFAR100 experiments with lower $P_{StatMix}$ probability (between $0.1$ and $0.4$) achieve better final accuracy. A possible explanation is that CIFAR100 is a more complex dataset and introducing too much noise through the \emph{StatMix} augmentation is no longer beneficial. This leads to a conclusion that the results on CIFAR100 could potentially be further optimized by decreasing the probability of \emph{StatMix} application.

\section{Concluding remarks}
In this work, \emph{StatMix}, a novel DA method designed for FL, has been introduced. \emph{StatMix} exchanges high level image statistics (two values per color channel). As a result, data privacy remains protected. At the same time, it has been empirically validated that using this method improves model accuracy, over baseline training (with no use of \emph{StatMix}), for two standard benchmark datasets, and two popular CNN architectures. Furthermore, \emph{StatMix} improves performance in classical, non-FL setup where the method helped in majority of cases. 

While application of \emph{StatMix} demonstrates very promising results, future work, 
aimed at verifying if sharing additional statistics (e.g. those related to hidden layers of the trained networks) could be beneficial. 
However, such an approach would be more expensive, when it comes to computation, since it would require local networks (in each node) to be trained at least twice. The first training would be needed to calculate statistics of the image in the inference phase in selected hidden layers of the network. 
These hidden-layer statistics could be then distributed to all nodes, and used in the process of final models training, similarly to the current \emph{StatMix} specification. Verification of this approach is planned as the next step in \emph{StatMix} development. 

\subsubsection{Acknowledgements} 
Research funded in part by the Centre for Priority Research Area Artificial Intelligence and Robotics of Warsaw University of Technology within the Excellence Initiative: Research University (IDUB) programme.

\bibliographystyle{splncs04}
\bibliography{manuscript}

\begin{thebibliography}{10}
\providecommand{\url}[1]{\texttt{#1}}
\providecommand{\urlprefix}{URL }
\providecommand{\doi}[1]{https://doi.org/#1}

\bibitem{non-IID}
Danilenka, A., Ganzha, M., Paprzycki, M., Ma{\'n}dziuk, J.: Using adversarial
  images to improve outcomes of federated learning for non-iid data. CoRR
  \textbf{abs/2206.08124} (2022)

\bibitem{IMAGENET}
Deng, J., Dong, W., Socher, R., Li, L.J., Li, K., Fei-Fei, L.: {ImageNet: A
  Large-Scale Hierarchical Image Database}. In: CVPR09 (2009)

\bibitem{FedLearnDifferentialPrivacy}
Geyer, R.C., Klein, T., Nabi, M.: Differentially private federated learning:
  {A} client level perspective. CoRR  \textbf{abs/1712.07557} (2017)

\bibitem{ZSLFederatedAugmentation}
Hao, W., El-Khamy, M., Lee, J., Zhang, J., Liang, K.J., Chen, C., Duke, L.C.:
  Towards fair federated learning with zero-shot data augmentation. In:
  Proceedings of the IEEE/CVF Conference on Computer Vision and Pattern
  Recognition (CVPR) Workshops. pp. 3310--3319 (June 2021)

\bibitem{PreActResNet18}
He, K., Zhang, X., Ren, S., Sun, J.: Identity mappings in deep residual
  networks. In: Leibe, B., Matas, J., Sebe, N., Welling, M. (eds.) Computer
  Vision - {ECCV} 2016 - 14th European Conference, Amsterdam, The Netherlands,
  October 11-14, 2016, Proceedings, Part {IV}. Lecture Notes in Computer
  Science, vol.~9908, pp. 630--645. Springer (2016).
  \doi{10.1007/978-3-319-46493-0\_38}

\bibitem{FedLearnEfficiency2}
Hsieh, K., Phanishayee, A., Mutlu, O., Gibbons, P.B.: The non-iid data quagmire
  of decentralized machine learning. In: Proceedings of the 37th International
  Conference on Machine Learning, {ICML} 2020, 13-18 July 2020, Virtual Event.
  Proceedings of Machine Learning Research, vol.~119, pp. 4387--4398. {PMLR}
  (2020)

\bibitem{GANFederatedAugmentation}
Jeong, E., Oh, S., Kim, H., Park, J., Bennis, M., Kim, S.:
  Communication-efficient on-device machine learning: Federated distillation
  and augmentation under non-iid private data. CoRR  \textbf{abs/1811.11479}
  (2018)

\bibitem{GANFederatedAugmentationv2}
Jeong, E., Oh, S., Park, J., Kim, H., Bennis, M., Kim, S.L.: Hiding in the
  crowd: Federated data augmentation for on-device learning. IEEE Intelligent
  Systems  \textbf{36}(5),  80--87 (2021). \doi{10.1109/MIS.2020.3028613}

\bibitem{FedLearnCanonical}
Kone{\v{c}}n{\'y}, J., McMahan, H.B., Ramage, D., Richt{\'{a}}rik, P.:
  Federated optimization: Distributed machine learning for on-device
  intelligence. CoRR  \textbf{abs/1610.02527} (2016)

\bibitem{FedLearnEfficiency}
Kone{\v{c}}n{\'y}, J., McMahan, H.B., Yu, F.X., Richt{\'{a}}rik, P., Suresh,
  A.T., Bacon, D.: Federated learning: Strategies for improving communication
  efficiency. CoRR  \textbf{abs/1610.05492} (2016)

\bibitem{cifar_10}
Krizhevsky, A., Nair, V., Hinton, G.: {CIFAR-10 (Canadian Institute for
  Advanced Research)}  (2009), \url{http://www.cs.toronto.edu/~kriz/cifar.html}

\bibitem{cifar_100}
Krizhevsky, A., Nair, V., Hinton, G.: {CIFAR-100 (Canadian Institute for
  Advanced Research)}  (2009), \url{http://www.cs.toronto.edu/~kriz/cifar.html}

\bibitem{AlexNet}
Krizhevsky, A., Sutskever, I., Hinton, G.E.: Imagenet classification with deep
  convolutional neural networks. In: Bartlett, P.L., Pereira, F.C.N., Burges,
  C.J.C., Bottou, L., Weinberger, K.Q. (eds.) Advances in Neural Information
  Processing Systems 25: 26th Annual Conference on Neural Information
  Processing Systems 2012. Proceedings of a meeting held December 3-6, 2012,
  Lake Tahoe, Nevada, United States. pp. 1106--1114 (2012),
  \url{https://proceedings.neurips.cc/paper/2012/hash/c399862d3b9d6b76c8436e924a68c45b-Abstract.html}

\bibitem{naszeSurvey}
Lewy, D., Ma{\'n}dziuk, J.: An overview of mixing augmentation methods and
  augmentation strategies. Artificial Intelligence Review  (2022).
  \doi{10.1007/s10462-022-10227-z}, published online 2022/06/30

\bibitem{Survey_2019}
Li, Q., Wen, Z., Wu, Z., Hu, S., Wang, N., Liu, X., He, B.: A survey on
  federated learning systems: Vision, hype and reality for data privacy and
  protection. CoRR  \textbf{abs/1907.09693} (2019)

\bibitem{FedProx}
Li, T., Sahu, A.K., Zaheer, M., Sanjabi, M., Talwalkar, A., Smith, V.:
  Federated optimization in heterogeneous networks. In: Dhillon, I.S.,
  Papailiopoulos, D.S., Sze, V. (eds.) Proceedings of Machine Learning and
  Systems 2020, MLSys 2020, Austin, TX, USA, March 2-4, 2020. mlsys.org (2020)

\bibitem{COCO}
Lin, T., Maire, M., Belongie, S.J., Hays, J., Perona, P., Ramanan, D.,
  Doll{\'{a}}r, P., Zitnick, C.L.: {Microsoft {{COCO:}} Common Objects in
  Context}. In: Fleet, D.J., Pajdla, T., Schiele, B., Tuytelaars, T. (eds.)
  {Computer Vision - {{ECCV}} 2014 - 13th European Conference, Zurich,
  Switzerland, September 6-12, 2014, Proceedings, Part {{V}}}. Lecture Notes in
  Computer Science, vol.~8693, pp. 740--755. Springer (2014).
  \doi{10.1007/978-3-319-10602-1\_48}

\bibitem{cosineannealing}
Loshchilov, I., Hutter, F.: {SGDR:} stochastic gradient descent with warm
  restarts. In: 5th International Conference on Learning Representations,
  {ICLR} 2017, Toulon, France, April 24-26, 2017, Conference Track Proceedings.
  OpenReview.net (2017), \url{https://openreview.net/forum?id=Skq89Scxx}

\bibitem{FedAvg}
McMahan, B., Moore, E., Ramage, D., Hampson, S., y~Arcas, B.A.:
  Communication-efficient learning of deep networks from decentralized data.
  In: Singh, A., Zhu, X.J. (eds.) Proceedings of the 20th International
  Conference on Artificial Intelligence and Statistics, {AISTATS} 2017, 20-22
  April 2017, Fort Lauderdale, FL, {USA}. Proceedings of Machine Learning
  Research, vol.~54, pp. 1273--1282. {PMLR} (2017)

\bibitem{FedLearnMixupUtilization}
Oh, S., Park, J., Jeong, E., Kim, H., Bennis, M., Kim, S.: Mix2fld: Downlink
  federated learning after uplink federated distillation with two-way mixup.
  {IEEE} Commun. Lett.  \textbf{24}(10),  2211--2215 (2020)

\bibitem{SGD}
Ruder, S.: An overview of gradient descent optimization algorithms. arXiv
  preprint arXiv:1609.04747  (2016)

\bibitem{FedLearnMixupUtilization2}
Shin, M., Hwang, C., Kim, J., Park, J., Bennis, M., Kim, S.: {XOR} mixup:
  Privacy-preserving data augmentation for one-shot federated learning. CoRR
  \textbf{abs/2006.05148} (2020)

\bibitem{FedMix}
Yoon, T., Shin, S., Hwang, S.J., Yang, E.: Fedmix: Approximation of mixup under
  mean augmented federated learning. In: 9th International Conference on
  Learning Representations, {ICLR} 2021, Virtual Event, Austria, May 3-7, 2021.
  OpenReview.net (2021)

\bibitem{DLA}
Yu, F., Wang, D., Shelhamer, E., Darrell, T.: Deep layer aggregation. In: 2018
  {IEEE} Conference on Computer Vision and Pattern Recognition, {CVPR} 2018,
  Salt Lake City, UT, USA, June 18-22, 2018. pp. 2403--2412. Computer Vision
  Foundation / {IEEE} Computer Society (2018). \doi{10.1109/CVPR.2018.00255}

\bibitem{FedLearnNNBlending}
Yurochkin, M., Agarwal, M., Ghosh, S., Greenewald, K.H., Hoang, T.N., Khazaeni,
  Y.: Bayesian nonparametric federated learning of neural networks. In:
  Chaudhuri, K., Salakhutdinov, R. (eds.) Proceedings of the 36th International
  Conference on Machine Learning, {ICML} 2019, 9-15 June 2019, Long Beach,
  California, {USA}. Proceedings of Machine Learning Research, vol.~97, pp.
  7252--7261. {PMLR} (2019)

\bibitem{mixup}
Zhang, H., Ciss{\'{e}}, M., Dauphin, Y.N., Lopez{-}Paz, D.: mixup: Beyond
  empirical risk minimization. In: 6th International Conference on Learning
  Representations, {ICLR} 2018, Vancouver, BC, Canada, April 30 - May 3, 2018,
  Conference Track Proceedings. OpenReview.net (2018)

\bibitem{FedLearnTrees}
Zhao, L., Ni, L., Hu, S., Chen, Y., Zhou, P., Xiao, F., Wu, L.: Inprivate
  digging: Enabling tree-based distributed data mining with differential
  privacy. In: 2018 {IEEE} Conference on Computer Communications, {INFOCOM}
  2018, Honolulu, HI, USA, April 16-19, 2018. pp. 2087--2095. {IEEE} (2018).
  \doi{10.1109/INFOCOM.2018.8486352}

\bibitem{FedLearnNonIID}
Zhao, Y., Li, M., Lai, L., Suda, N., Civin, D., Chandra, V.: Federated learning
  with non-iid data. CoRR  \textbf{abs/1806.00582} (2018)

\bibitem{mix_style}
Zhou, K., Yang, Y., Qiao, Y., Xiang, T.: Domain generalization with mixstyle.
  CoRR  \textbf{abs/2104.02008} (2021)

\bibitem{FedLearnVertical}
Zhuang~Yan, Li~Guoliang, F.J.: A survey on entity alignment of knowledge base.
  Journal of Computer Research and Development  \textbf{53}(1), 165 (2016).
  \doi{10.7544/issn1000-1239.2016.20150661}

\end{thebibliography}
\end{document}